# Lighter Stacked Hourglass Human Pose Estimation

Ahmed Elhagry
20020054@mbzuai.ac.ae

Mohamed Saeed
20020065@mbzuai.ac.ae

Musie Araia
20020058@mbzuai.ac.ae

## 1. Abstract

Human pose estimation (HPE) is one of the most challenging tasks in computer vision as humans are deformable by nature and thus their pose has so much variance. HPE aims to correctly identify the main joint locations of a single person or multiple people in a given image or video. Locating joints of a person in images or videos is an important task that can be applied in action recognition and object tracking. As have many computer vision tasks, HPE has advanced massively with the introduction of deep learning to the field. In this paper, we focus on one of the deep learning-based approaches of HPE proposed by Newell et al. [1], which they named the stacked hourglass network. Their approach is widely used in many applications and is regarded as one of the best works in this area.

The main focus of their approach is to capture as much information as it can at all possible scales so that a coherent understanding of the local features and full-body location is achieved. Their findings demonstrate that important cues such as orientation of a person, arrangement of limbs, and adjacent joints' relative location can be identified from multiple scales at different resolutions. To do so, they makes use of a single pipeline to process images in multiple resolutions, which comprises a skip layer to not lose spatial information at each resolution. The resolution of the images stretches as lower as 4x4 to make sure that a smaller spatial feature is included.

In this study, we study the effect of architectural modifications on the computational speed and accuracy of the network.

## 2. Architecture Design

The original architecture is composed of stacked hourglasses, which are each a combinations of various convolution and max-pooling layers designed to lower down the resolution of the image and capture some features in the way. Then unsampling and combination of features at all scales is carried out right after the image reaches the lowest resolution. This process is called top-down. In addition, unsampling of the nearest neighbor is done and is added to the combined feature elementwise so that information from the two adjacent resolutions is captured. The hourglass also includes a bottom-up process, which is symmetric to the top-down. Finally, two 1x1 convolutions are applied to the output to get the final prediction. The output is given in the form of a heatmap, which represents the probability of occurrence of joints pixel-wise.

As shown in Fig. 1, the entire architecture comprises multiple hourglasses stacked next to each other where the output of one is input to the next one. This network of hourglasses helps to evaluate features from top-down to bottom-up repeatedly throughout the entire image. This repeated top-down and bottom-up processing helps the network capture high-level features at an early stage and a deeper evaluation of the features takes place at later stages. It also preserves the spatial location of features so that the network localizes joints effectively.

It has been noted that the overall performance of the hourglass can be improved by applying multiple smaller filters instead of a larger one, for example, using two 3x3 filters as opposed to a 5x5 filter. Furthermore, convolving with a 1x1 filter to reduce the resolution also enhances its performance [2]. Hence, this architecture has applied 3x3 or less sized filters throughout. The network shows an improvement when a residual learning module is applied instead of standard convolutional layers. Also, it is worth noting that 64x64 input images were provided to the network instead of high-resolution images to avoid extensive usage of GPU memory. This has no negative effect on the performance of its prediction[3].

The performance of this architecture is evaluated on MPII Human pose dataset, which contains around 25k images in which 40k people are annotated. The images in MPII are captured from various human activities, and thus diverse challenging human poses are included. In case there are multiple people in a given image, the network was designed to label a person in the center. In MPII, each image is annotated with its scale and center so locating the center is not a difficult task. After locating a person at the center, the image is cropped to only include the target person and then resized to 256x256 pixels. Finally, data augmentations such as scaling, and rotation are being applied to the image.

The baseline architecture consists of eight stacked hour-glasses, each encompassing a single residual module at every scale. They experiment with the effect of the number of residual modules in different network architectures. For example, 4 stacked hourglasses with 2 residual modules versus 2 stacked hourglasses with 4 residual modules. After comprehensive comparison, they found out that stacked hourglasses outperform a single hourglass when tested with the same number of layers and parameters. Furthermore, the effect of applying supervision at different stages of the network was compared, and the result shows supervision is more effective when applied at the two resolutions prior to the final output resolution right after upsampling. In this study, we conduct the following experiments to explore potential improvements on performance and training time.

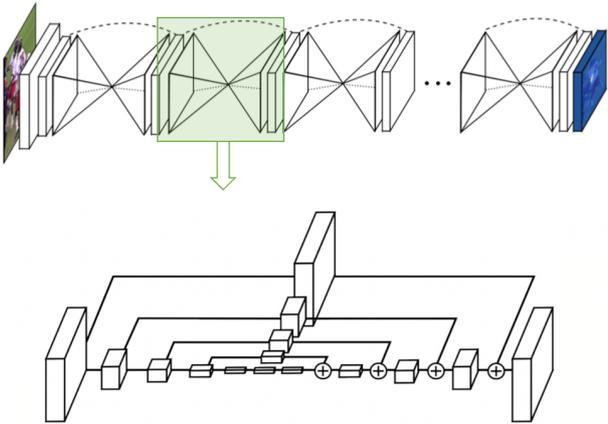

Figure 1: Repeated bottom-up and top-down carried away by stack of hourglasses with residual modules.

### 2.1. Depthwise Separable Convolution

Depthwise separable convolutions consist of a depthwise convolution, in which an independent kernel is used for each channel, followed by a pointwise (1x1) convolution with a kernel that has a depth equal to the number of channels. This greatly decreases the number of parameters and hence the computational cost, at the expense of a reduction in accuracy [4].

### 2.2. Dilated Convolution

Dilated convolution is another convolution-based modification to the network architecture. The main purpose of a dilated convolution is to increase the receptive field without increasing the computational cost much. This is done by applying the kernel to an input with gaps that are created by skipping some pixels. Relative to standard convolutions, dilated convolutions have the same or lower number of parameters, depending on the dilation scale, but result in a wider receptive field [4].

### 2.3. Ghost Bottleneck

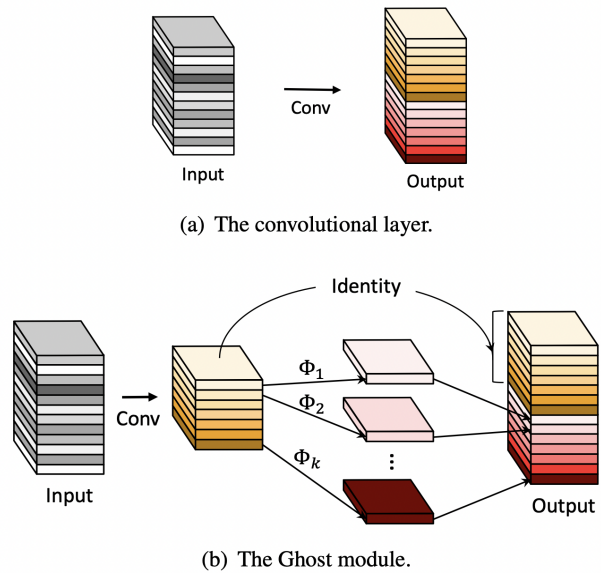

Figure 2: The Ghost Module introduced by [5] produces more feature maps than a standard convolution layer using cheap linear operations.

Han et al. [5] proposed a residual network structure that utilizes linear operations that are computationally cheaper than convolution to produce feature maps. A subset of intrinsic feature maps generated by standard convolution are then used to generate more "ghost" feature maps as can be seen in Fig. 2. A bottleneck structure (Ghost bottleneck) (Fig. 3d) was designed based on the aforementioned Ghost module. In this study, the Ghost bottleneck was used to replace the original bottleneck in two of the experiments.

## 3. Implementation

The implementation of the stacked hourglass network was based on the pytorch-stacked-hourglass GitHub repository by anibali. Their original implementation of the eight-stack hourglass architecture was used as a baseline for this study. We modified their code to implement each of the modifications listed below and compare the results with the baseline models.

### 3.1. Modifications on Stacked Hourglass Network

Modifications were made to the architecture of the original network to reduce computational complexity with minimal effect on the accuracy. Hyperparameters were kept constant across all the experiments to allow for a fair comparison. A learning rate of 5e-4, RMSProp optimizer, 220 training epochs and a batch size of 6 (training and testing) were used throughout. Training was done using a Quadro RTX

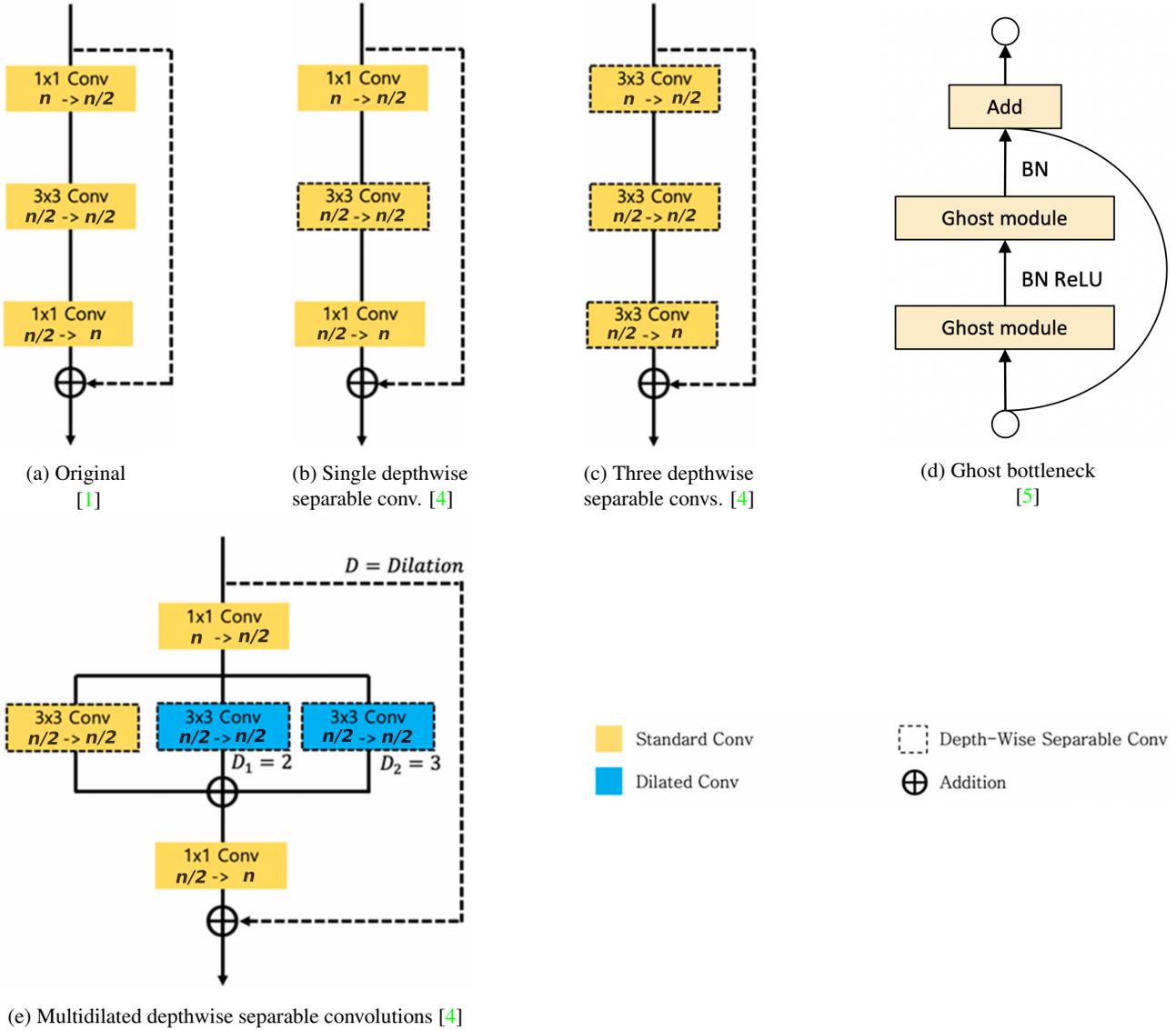

Figure 3: Summary of implemented bottleneck modifications **(a)** Original bottleneck architecture [1][4] **(b)** Modified bottleneck with single depthwise separable convolution [4] **(c)** Modified bottleneck with three depthwise separable convolutions [4] **(d)** Ghost bottleneck based on Ghost modules [5] **(e)** Modified bottleneck with three depthwise separated convolutions at different dilation scales [4]

6000 GPU with 24 workers running on an Intel Xeon Silver 4215 CPU. Apart from changing the number of stacks in the original network, the architectural changes made were focused on the bottleneck and included the introduction of depthwise separable convolutions, dilated convolutions and replacement of the original bottlenecks with Ghost bottlenecks. The modifications were implemented either on top of the base single hourglass architecture, two-stacked hourglass architecture or both. The details of each are in the following subsections. Unless otherwise specified, the number of input feature maps for the first two bottlenecks is 64, and 128 for the remaining. Most of the changes were made based on previous findings by [4] and [5].

### 3.1.1 Single Depthwise Separable Convolution

Two experiments were conducted by changing the second convolutional layer in the bottleneck to a depthwise separable convolution (Fig. 3b). This modification was tested with the single-stack and two-stack hourglass architectures.

| Method | Validation Accuracy (%) | | | | | | | | Num. of Parameters | Time | |
|---|---|---|---|---|---|---|---|---|---|---|---|
| | Head | Shoulder | Elbow | Wrist | Hip | Knee | Ankle | Mean | | Training (Hrs) | Inference (Mins) |
| Baseline | | | | | | | | | | | |
| 8-Stack Hourglass | 95.50 | 92.92 | 87.78 | 83.76 | 82.57 | 81.04 | 79.00 | 86.23 | 97.7M | 97:34 | 01:51 |
| Original Architecture with Different Number of Stacks | | | | | | | | | | | |
| Single Hourglass | 94.61 | 92.07 | 84.56 | 78.78 | 80.96 | 75.44 | 71.26 | 82.71 | 12.6M | 33:11 | 01:53 |
| 2-Stack Hourglass | 95.05 | 93.27 | 87.06 | 82.13 | **83.45** | 80.07 | 76.10 | 85.47 | 24.8M | 37:01 | 01:48 |
| Modified Single Hourglass | | | | | | | | | | | |
| 1 Depthwise Separable | 94.17 | 91.47 | 82.50 | 76.91 | 79.89 | 74.71 | 69.46 | 81.49 | 5.0M | 38.31 | 00:58 |
| 3 Depthwise Separable | 94.44 | 92.87 | 84.27 | 78.55 | 81.46 | 76.97 | 71.75 | 83.10 | 5.4M | **27:30** | 00:58 |
| Ghost Bottleneck | 93.25 | 89.84 | 79.44 | 73.09 | 77.36 | 70.97 | 64.10 | 78.53 | 2.2M | 31:43 | **00:57** |
| Ghost + Reduced Features | 93.89 | 89.49 | 78.97 | 72.66 | 77.03 | 70.64 | 62.02 | 78.02 | **2.1M** | **27:30** | 00:59 |
| Multidilated | 94.78 | 92.58 | 85.07 | 79.75 | 80.70 | 77.35 | 72.70 | 83.46 | 13.7M | 33:26 | 01:00 |
| Modified 2-Stack Hourglass | | | | | | | | | | | |
| 1 Depthwise Separable | 95.02 | 92.80 | 85.84 | 80.64 | 82.08 | 78.54 | 74.66 | 84.39 | 9.9M | 42:31 | 00:58 |
| Multidilated | **95.57** | **93.46** | **87.63** | **82.44** | 82.92 | 79.85 | **76.78** | **85.64** | 26.9M | 44:28 | 00:59 |

Table 1: Summary of results - training times, inference times and final validation PChk scores using the 493 validation images from the MPII dataset

### 3.1.2 Three Depthwise Separable Convolutions

A single experiment was conducted with the three convolutional layers in the bottleneck being replaced by 3x3 depthwise separable convolutions (Fig. 3c). This change was tested on the single stack architecture only.

### 3.1.3 Ghost Bottleneck

One experiment was done by replacing the original bottleneck on the single stack architecture with a Ghost bottleneck (Fig. 3d).

### 3.1.4 Ghost Bottleneck with Reduced Features

Another experiment was done using the Ghost bottleneck (Fig. 3d) with a reduction of the number of input feature maps of the first two bottlenecks in the residual block of the hourglass to 32 only rather than the 64 used by the remaining experiments.

### 3.1.5 Multidilated Depthwise Separable Convolutions

The final modification was the replacement of the second convolution in the bottleneck with 3 parallel depthwise separable convolutions with different dilation scales (1, 2 and 3) (Fig. 3e). This was tested on both the single and two stack architectures.

## 4. Results and Discussion

Based on the results (Table 1), the multidialted two-stack hourglass architecture was the highest in the mean validation accuracy achieving 85.64%. It achieved the highest accuracy locating most of the body parts, while being slightly outperformed on two joints by the original two-stack hourglass model. However, the training time was comparatively high. The high accuracy is because of the multidilated light residual block being a building block, that widens the convolutional receptive field. This has an impact on learning the features detected on the scale of the full human body. Besides, the training time was the longest because of the relative complexity of the multidilated block, especially on the two-stack hourglass model. Depthwise separable convolutions increased the validation accuracy of both the single and the two-stack hourglass remarkably, while decreasing the number of parameters and computation speed. In addition, a modification made by the ghost bottleneck was the lowest in the mean validation accuracy, as the simple ghost module creates intrinsic feature maps from fewer convolutions and then creates additional "ghost" feature maps using linear transformations. Overall, the multidilated two-stack hourglass model with depthwise separable convolutions is the best balance between time and accuracy in our opinion, achieving a 54% reduction in training time, 47% reduction in inference time and only a 0.77% reduction in mean accuracy compared to the baseline.

## 5. Conclusion

Computer vision applications may not always prioritize getting the best possible accuracy from the model. In many real-life scenarios, other factors are essential too including the training cost (computational complexity and resources), inference time and the flexibility of the model to be customized and developed. In applications like human pose estimation, all of these factors are necessary. In fact, it is a critical application when it comes to real-time sports analysis and health.